# Control of nonlinear, complex and black-boxed greenhouse system with reinforcement learning

Byunghyun Ban[1], Soobin Kim[2]

[1]Korea Advanced Institute of Science and Technology and
[2]Kyungbuk National University
[1]Daejun and [2]Daegu, Republic of Korea
{bhban@kaist.ac.kr, shastar5@knu.ac.kr}

*Abstract*— **Modern control theories such as systems engineering approaches try to solve nonlinear system problems by revelation of causal relationship or co-relationship among the components; most of those approaches focus on control of sophisticatedly modeled white-boxed systems. We suggest an application of actor-critic reinforcement learning approach to control a nonlinear, complex and black-boxed system. We demonstrated this approach on artificial green-house environment simulator all of whose control inputs have several side effects so human cannot figure out how to control this system easily. Our approach succeeded to maintain the circumstance at least 20 times longer than PID and Deep Q Learning.**

*Keyword*s—**System Engineering, Deep Learning, Machine Learning, Reinforcement Learning**

## I. INTRODUCTION

Frequently, the approaches in nonlinear system control is to model the target system in highly sophisticated level. For example, systems biologist starts with defining the components of the system; the number of components are usually tens but sometimes thousands. The next step is to revealing interaction among all components and to draw interactional network. Continuous scaled systems are usually modeled in ordinary differential equation model [2][3]. When the number of component is too great to handle, researchers apply discrete model such as Boolean networks and write full truth table for individual interactions or establish Boolean equations for each component's input condition [4]. After modeling the system, they apply complicated control methods [5][6][7][8], which usually have non-polynomial computational cost.

Those routine is time-consuming and not realistic in two points. First, when the model is big enough, a researcher may fail to gather all interactions among the components because current approaches mainly depends on paper search. Second, control methods for nonlinear systems often takes exponential time complexity so applying them on large scaled system is not feasible. Not to mention most of the methods are not even applicable when the model is black-box status.

The idea of our work is to provide a control method for black-boxed complex system whose input and output are only observable variables.

## II. REINFORCEMENT LEARNING BACKGROUNDS

Reinforcement learning is interactions between the environment and agent over some discrete time steps, where the agent learns how to apply action on environment to maximize future rewards. At each time step t, the agent observes current state $s_t$ and decide an action $a_t$ according to its policy $\pi$ which, is a function whose domain is $s_t$ and has possible actions as codomain. The model decides future action by approximation of action value which is expected future rewards by applying $a_t$ to the environment.

### A. Critic-only methods

Critic-only methods, value based reinforcement learning methods in other words, aims to approximate the value or action-value function Q (s, a; w), where $w$ is internal parameter for Q function [9] and is updated with various reinforcement learning algorithms [10]. For example, in one-step Q-learning [11], the learning is done by minimizing the loss function;

$$L_t(w_t) = \mathbb{E}\left(r + \gamma \max_{a'} Q(s', a'; w_{t-1}) - Q(s, a, w_t)\right)^2,$$

where $s'$ is the state encountered after state $s$ and $\gamma$ is discount factor. The output value of value-based reinforcement learning agent neural network is Q.

### B. Actor-only methods

Policy gradient methods belongs to actor-only methods. The output of policy-based reinforcement learning agent neural network is the policy function $\pi_\theta(s, a)$, not Q itself. [12] Policy based reinforcement learning is to find best $\theta$ for $\pi_\theta$. To measure the quality of $\pi_\theta$, policy based approaches suggest various objective functions. By the policy gradient theorem proposed by Sutton et al., the policy gradient is described as

$$\sum_s \sum_{t=0}^\infty \gamma^t \Pr\{s_t = s|s_0, \pi\} \sum_a \frac{\partial \pi(s,a)}{\partial \theta} Q^\pi(s, a).$$

### C. Actor-critic policy gradient

Actor-critic methods aims to combine and maximize the strong point of value-based and policy-based models. This approach train 2 models, critic model and actor model, at the same time.

Thanks to 상상텃밭 Inc.

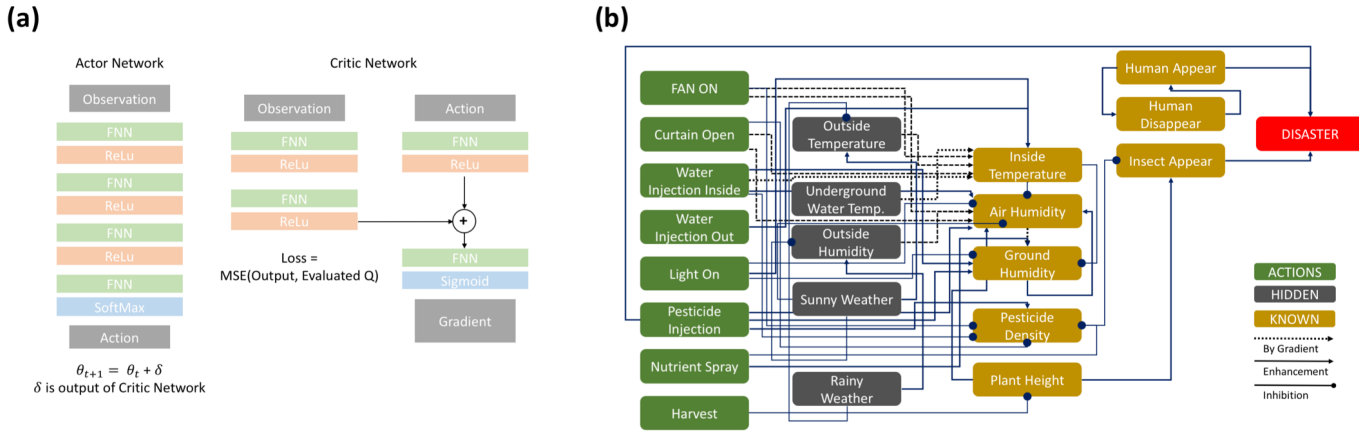

[Figure 1] **(a) Network Architecture** The reinforcement learning agent consists of two independent deep neural networks. Actor network (left) makes decision on which action to make at every time step and is updated with gradient provided by critic network. Critic network (right) receives state of the environment and action of actor network at the same time. Critic network has two networks for observation and action, and addition of each layer's output passes through single FNN layer. The output of critic network is the gradient for actor network. **(b) Environment** The network describes interaction of variables which consists of or affects the greenhouse system. It has 8 possible actions, 5 hidden variables and 7 observable parameters. Although human appearance and disappearance are described as independent component, it can be described as single Boolean variable in mathematical expressions. Solid arrow means enhancement or activation, and dot-headed arrow means inhibition. Dashed line means this interaction can work as both enhancement and inhibition, depending on their gradient.

Critic is a temporal difference learning algorithm with a linearly parameterized approximation architecture for action-value function Q. [13] With linear approximation, with the basis function $\emptyset$, the value function is described as $Q_w(s,a) = \emptyset(s,a)^T w$. [14][15] It updates action-value function parameter w by linear TD. Critic estimates current policy with Q function. Actor updates policy parameter $\theta$, in direction suggested by critic.

Actor-critic algorithms follow an approximate policy gradient:

$$\nabla_\theta J(\theta) \approx \mathbb{E}[\nabla_\theta \log \pi_\theta(s,a) Q_w(s,a)],$$
$$\Delta\theta = \alpha \nabla_\theta \log \pi_\theta(s,a) Q_w(s,a). \ [14].$$

## III. METHOD

Our approach is implementation of actor-critic policy gradient based reinforcement learning agent with artificial neural networks. This deep reinforcement learning algorithm tries to find out how to maximize the future rewards: maximize control performance.

### A. Network Architecture

The agent is combined actor-critic networks. The structural backbone is described on Figure 1 (a). The structure is same as pemami4911's work on his GitHub repository [16] but parameters are changed to fit for nonlinear, black-boxed and complex system control.

Actor network has 4 layers of fully connected neural networks. The activation functions are ReLu except the final layer; it is softmax function. The first and second layers has 128 neurons while the third layer has 32 neurons. The number of the neurons in final layer is same with the number of possible actions. The actor produces one-hot vector formed output corresponding with an action to be applied to the environment. Input to the actor is observation only. Network parameters are updated with critic output value.

Critic network has slightly unique structure. It is 3-layered FNN model, which has fuse operation on the second layer. Another single layer receives actor's output to be merged with encoded observation information. The backpropagation of critic network follows both subnetworks at the same time so this network naturally trains how to optimize the actor policy by evaluating actor policy regarding the observation at the same time.

Network parameters are updated to reduce the mean squared error between previously predicted Q value with discounted real rewards.

### B. Training

For each step, the agent receives observation form the environment. Actor decides which action to make regarding the observed state. For each step, the critic updates its internal parameter for Q function and returns gradient for actor. Then the actor is updated with the given gradient. This routine is repeated until the environment returns game over signal; after an episode is done, new episode starts with cleared environment.

## IV. EXPERIMENT

### A. Environment

The environment used for experiment is artificial greenhouse simulator. It simulates a greenhouse on very barren environment; weather changes a lot and sunlight is too strong so its inside circumstance becomes detrimental for the plants only in 10 hours. The derivatives of internal variables are proportional to difference between the internal value and external value and the coefficient for the derivative is chosen stochastically between [0.01, 0.2]. Perturbation on the external weather and entrance of human user randomly occurs.

The Environment has 7 observable values; current temperature, current humidity of air, current humidity of ground, current height of the plants, pesticide density, the existence of human and insects, and receives 8 actions; fan on, curtain open, water injection inside, water injection outside, pesticide injection, light on, spraying nutrients and harvest.

Figure 1 (b) shows the complexity of greenhouse system. Each action affects several internal values at the same time so any action can cause undesirable effects on the system. For example, if an agent take 'curtain open' action to reduce internal temperature, current humidity of both air and ground is changed with gradient calculated with outside values, and the pesticide density drops too.

Usually, Smart-farm owners apply several PIDs to maintain each value individually. However, it is very inefficient; notwithstanding the initial cost, it consumes lots of energies; it's common for them to activate the air conditioner, dehumidifier and humidifier at the same time to maintain all the variable at the same time with individual PIDs. The agent of our method learned how to avoid such problem while maintaining the environment appropriately, taking least costs. If the environment is still sustainable after each hour, it returns +1 reward; total reward is hours for which the agent can maintain the environment.

Details on the environment are described on the supplementary information.

*B. Training*

The agent only observes current status but never receives information on interactions of internal variables to figure out appropriate control input. The environment is initialized as follows; allowed temperature range as [10, 100], allowed air humidity as [10, 90], allowed ground humidity as [30, 80], initial temperature as 21, initial air humidity as 50, initial ground humidity as 50. The episode is interrupted even a single variable exceeds the range.

To avoid converging in local minimum, actor's sample action is ignored with probability eps, which is initially 0.95 and decays to 10% per 9 episodes; initially the model train and calculates value and policy for random action. For each step, the action and rewards are stored in a queue and 256 samples are drawn for mini batch.

*C. Computational Environments*

The agent was implemented with Tensor Flow on python 3. Numpy library is used for training routine and the implementation of environment. NVIDIA GTX 1080 8gb was used for training.

|  | Scores | | |
| --- | --- | --- | --- |
|  | *Trial A* | *Trial B* | *Trial C* |
| Mean | 6.02 | 6.01 | 10.04 |
| Stddev | 1.84 | 1.80 | 1.80 |

**[Table 1] PID Experiment Results**

V. RESULTS

We compared actor-critic network with conventional PID and Deep Q learning algorithm. Deep reinforcement learning approach showed better performance than conventional PID approach.

*A. Conventional PID*

PID controller measures each observable variable, and apply actions which can draw it into normal state. We applied it with 3 different conditions.

      Trial A: Apply actions at every hour

      Trial B: Apply actions with 3-hour interval

      Trial C: Apply actions after $6^{th}$ hour.

Results from Trial A and B was frustrating. The environment itself requires average 10 hours to be destroyed without any control input but the result showed that PID accelerated destruction of the system. The trial C started after 5 hours passed, and managed to maintain the environment slightly longer than 10 hours. (Table 1)

It was mainly due to side effects of each control inputs. The major variable for the system is temperature. However, most control input related to temperature drops ground humidity dramatically so the system easily get dry.

*B. Deep Q learning*

Deep Q learning suggested by Google DeepMind [1], implemented with 4 layers of fully connected networks failed to achieve desirable performance. As the circumstance is sustainable up to 10 hours without any actions, this model definitely failed to learn how to maintain the greenhouse. It sometimes showed more than 10 scores but most of the trial showed less than 10 scores.

*C. Actor-critic policy gradient*

Our model showed outstanding result. After 2,000 episodes, the score increased dramatically and it succeeded to maintain the environment for 200~350 hours which is enough for farmers to grow and harvest lettuce. Regarding that the artificial greenhouse is unrealistically infertile, it can already be applied commercially in real world especially in the climate around the Republic of Korea.

VI. CONCLUSION

We showed the performance comparison between PID and simple DQN with actor-critic network on controlling so extreme environment even a single action changes several internal values of the system. Moreover, the system is nonlinear and observation is limited so the agent doesn't know the internal dynamics (black-boxed). It is not surprising that simple DQN, critic-only approach, failed to learn how to control such complex system.

Actor-critic policy gradient method showed surprisingly great performance because the complex rules are naturally trained not only on the action-value function but also on the

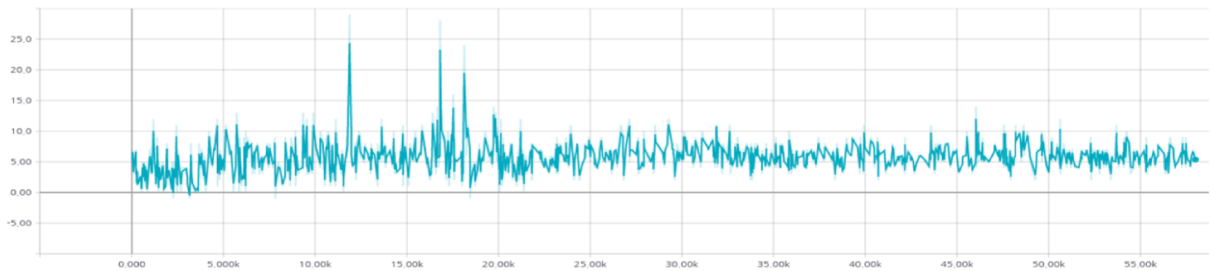

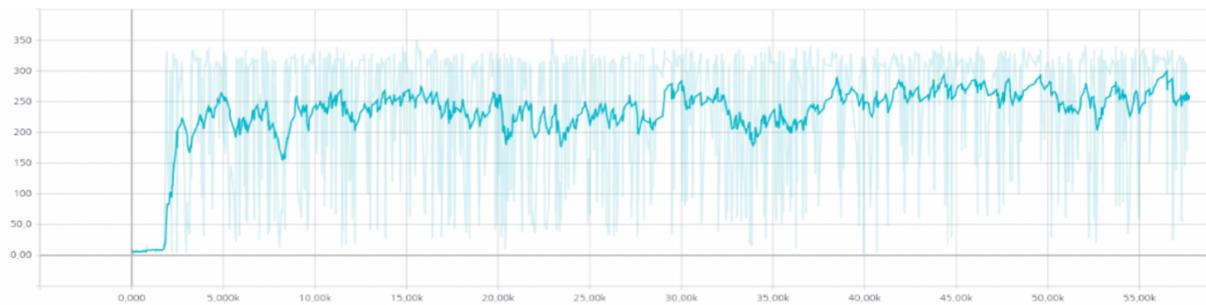

[Figure 2] Experiment result of each deep reinforcement learning algorithm. **(a) Deep Q learning** algorithm failed to achieve and maintain high-score. Its peak score was 28. **(b) Actor-critic policy gradient method** achieved much higher score than Deep Q learning. Although the variance is high, even minimum peaks were higher than DQN's score.

policy gradient so that the agent could to figure out the appropriate policy and future rewards at the same time. We suggest applying actor-critic policy gradient on complex system control problems.

## Supplementary Information

This section covers details of the greenhouse environment.

### A. Hidden Parameters

The environment has both observable and hidden parameters. For example, (a) outside temperature, (b) outside humidity, (c) weather, (d) temperature of the water are not observable to the agent. But they affect the internal state change.

### B. Observable Parameters

The environment Vinylhouse provides observation of the system status. An agent can observe (a) current temperature inside the greenhouse, (b) current humidity of inside air, (c) current humidity of the ground, (d) current height of plants inside, (e) current density of pesticide, (f) existence of a human inside the greenhouse, (g) existence of insect inside the greenhouse.

### C. Gradient

We supposed that the internal parameter of this nonlinear system is changed by simple different equation below;

$$\frac{\Delta s}{\Delta t} = \alpha(\varphi - s),$$

where s is single parameter such as current humidity and $\varphi$ is comparison partner. $\alpha$ is random float between [0.01, 0.2]; this value is randomly chosen for every step of gradient calculation to increase difficulty of maintenance problem.

The update for internal parameter can be described;

$$s_{t+1} = s_t + \Delta s_t, \text{ where}$$
$$\Delta s_t = \frac{\Delta s_t}{\Delta t} \Delta t = \alpha(\varphi - s_t)\Delta t.$$

As $\Delta t$ is just 1 for each discrete time step, update rule for internal parameter is simply approximated to $s_{t+1} \approx s_t + \alpha(\varphi - s_t)$.

### D. Actions

The environment may receive 8 actions. Each action is modeled to realistic ones. Usually, smart farm owners apply several PIDs to maintain each value individually. However, it is very inefficient; notwithstanding the initial cost, it consumes lots of energies; it's common for them to activate the air conditioner, dehumidifier and humidifier at the same time to maintain all the variable at the same time with individual PIDs. The agent will learn how to avoid such problem while maintaining the environment appropriately, taking least costs.

#### 1) Fan On

Open the window and activate ventilator. It continues for 10 hours. For each hour, current temperature, humidity of air, humidity of ground, density of pesticide changes. $T_t$ is inside temperature at time t and $H\_a_t$ is humidity of the air at time t and $P_t$ is pesticide density at time t.

$$T_{t+1} \approx T_t + \alpha(Outside\_temparature - T_t)$$
$$H\_a_{t+1} \approx H\_a_t + \alpha(Outside\_humidity - H\_a_t)$$
$$H\_g_{t+1} \approx H\_g_t + 0.1 \times \alpha(H\_a_t - H\_g_t)$$
$$P_{t+1} \approx 0.9 \times P_t$$

#### 2) Curtain Open

Open the side curtain of greenhouse. It continues for 30 hours. For each hour, current temperature, humidity of air, humidity of ground, density of pesticide changes. $T_t$ is inside temperature at time t and $H\_a_t$ is humidity of the air at time t and $P_t$ is pesticide density at time t.

$$T_{t+1} \approx T_t + 3 \times \alpha(Outside\_temparature - T_t)$$
$$H\_a_{t+1} \approx H\_a_t + 3 \times \alpha(Outside\_humidity - H\_a_t)$$
$$H\_g_{t+1} \approx H\_g_t + 0.2 \times \alpha(H\_a_t - H\_g_t)$$
$$P_{t+1} \approx 0.8 \times P_t$$

#### 3) Water Injection Inside

Switch on the sprinkler inside the greenhouse. It continues for 10 hours. For each hour, current temperature, humidity of air, humidity of ground, density of pesticide changes. $T_t$ is inside temperature at time t and $H\_a_t$ is humidity of the air at time t and $P_t$ is pesticide density at time t.

$$T_{t+1} \approx T_t + 3 \times \alpha(water\_temparature - T_t)$$
$$H\_a_{t+1} \approx 1.1 \times H\_a_t$$
$$H\_g_{t+1} \approx 1.3 \times H\_g_t$$
$$P_{t+1} \approx 0.7 \times P_t$$

#### 4) Water Injection Outside

Activate rooftop water flow. It continues for 20 hours. For each hour, current temperature, humidity of air, humidity of ground changes. $T_t$ is inside temperature at time t and $H\_a_t$ is humidity of the air at time.

$$T_{t+1} \approx T_t + 0.5$$
$$H\_a_{t+1} \approx 0.9 \times H\_a_t$$
$$H\_g_{t+1} \approx 0.95 \times H\_g_t$$

#### 5) Pesticide Injection

Spray pesticide inside the greenhouse. It continues for 10 hours. For each hour, humidity of air, humidity of ground, density of pesticide changes. $H\_a_t$ is humidity of the air at time t and $P_t$ is pesticide density at time t.

$$H\_a_{t+1} \approx 1.1 \times H\_a_t$$
$$H\_g_{t+1} \approx 1.2 \times H\_g_t$$
$$P_{t+1} \approx 2 \times P_t$$

#### 6) Light on

Turn on the plant growth stimulating light. It continues for 50 hours. For each hour, current temperature, humidity of air, humidity of ground changes. $T_t$ is inside temperature at time t and $H\_a_t$ is humidity of the air at time.

$$T_{t+1} \approx 1.01 \times T_t$$
$$H\_a_{t+1} \approx 0.95 \times H\_a_t$$
$$H\_g_{t+1} \approx 0.98 \times H\_g_t$$

#### 7) Nutrient Spray

Spray nutrient inside the greenhouse. It continues for 10 hours. For each hour, humidity of air, humidity of ground, density of pesticide changes. $H\_a_t$ is humidity of the air at time t and $P_t$ is pesticide density at time t.

$$H\_a_{t+1} \approx 1.1 \times H\_a_t$$
$$H\_g_{t+1} \approx 1.2 \times H\_g_t$$
$$P_{t+1} \approx 0.8 \times P_t$$

#### 8) Harvest

Doesn't change any internal variable. If plant height is not sufficient to be harvested or too tall, stop the episode. Otherwise, put huge reward.

*E. Perturbations*

The environment got perturbations by outside circumstance at every time step.

*1) By weather change*

Periodically the system decided whether to change the weather or not. If whether is changed, also outside temperature is changed too. Apply gradient on inside temperature.

*2) By rain*

If it's raining outside, set outside humidity value as maximum. Then apply gradient for inside air humidity. Also drop the temperature along the gradient.

*3) By sunny weather*

Reduce outside humidity into a random value among [10, 30], which is average humidity of south Korea. Applying gradient on temperature and increase its value +4. If temperature is increased, reduce current humidity among 0~6 random degrees. If current plant height is more than 50, apply inverse gradient on air humidity to reflect transpiration rate.

*4) Interaction between internal parameters*

Current humidity of ground is updated with gradient with current air humidity. Current plant height increases a little bit.

*5) Correction*

If any value exceeds 100, reduce it to 100. Also, any negative value is updated into 0.

*6) Human Entrance*

If human hasn't appeared for 30 hours, a human enters the greenhouse with 1/3 probability. If a human has existed for more than 20 hours, he disappears with 4/5 probability.

*7) Insect Attack*

If pesticide has been applied in 600 hours, insect doesn't appear. Otherwise, it appears with 1/10 probability.

*F. Interact between the agent*

Whenever an episode starts, the parameters are initialized. For each step, environment receives action from actor network and apply it. After that, a perturbation is applied to the environment too. Then the agent observes the environment.

*G. Terminal condition*

If any observable variable exceeds allowed range, the environment returns game over signal to agent and get ready for another episode. Also, the game may be finished by harvest action.